RESEARCH ARTICLE

# Enhancing Stability in Training Conditional Generative Adversarial Networks via Selective Data Matching

**KYEONGBO KONG [1], (Member, IEEE), KYUNGHUN KIM [2], AND SUK-JU KANG [3], (Member, IEEE)**
[1]Department of Electrical and Electronics Engineering, Pusan National University, Busan 46241, Republic of Korea
[2]NHN Cloud, Seongnam-si 13487, Republic of Korea
[3]Department of Electronic Engineering, Sogang University, Seoul 04017, Republic of Korea

Corresponding author: Suk-Ju Kang (sjkang@sogang.ac.kr)

This work was supported in part by the National Research Foundation of Korea (NRF) funded by Korean Government through the Ministry of Science and ICT (MSIT), South Korea, under Grant RS-2024-00414230; in part by the National Supercomputing Center with Supercomputing Resources, including Technical Support under Grant KSC-2023-CRE-0444; in part by MSIT under the Information Technology Research Center (ITRC) Support Program supervised by the Institute for Information Communications Technology Planning Evaluation (IITP) under Grant IITP-2024-RS-2023-00260091; and in part by NRF funded by the Ministry of Education under Grant 2020R1A6A3A01098940 and Grant 2021R1I1A1A01051225.

**ABSTRACT** Conditional generative adversarial networks (cGANs) have demonstrated remarkable success due to their class-wise controllability and superior quality for complex generation tasks. Typical cGANs solve the joint distribution matching problem by decomposing two easier sub-problems: marginal matching and conditional matching. In this paper, we proposes a simple but effective training methodology, selective focusing learning, which enforces the discriminator and generator to learn easy samples of each class rapidly while maintaining diversity. Our key idea is to selectively apply conditional and joint matching for the data in each mini-batch.Specifically, we first select the samples with the highest scores when sorted using the conditional term of the discriminator outputs (real and generated samples). Then we optimize the model using the selected samples with only conditional matching and the other samples with joint matching. From our toy experiments, we found that it is the best to apply only conditional matching to certain samples due to the content-aware optimization of the discriminator. We conducted experiments on ImageNet (64 $\times$ 64 and 128 $\times$ 128), CIFAR-10, CIFAR-100 datasets, and Mixture of Gaussian, noisy label settings to demonstrate that the proposed method can substantially (up to 35.18% in terms of FID) improve all indicators with 10 independent trials. Code is available at https://github.com/pnu-cvsp/Enhancing-Stability-in-Training-Conditional-GAN-via-Selective-Data-Matching.

## I. INTRODUCTION

Generative Adversarial Network (GAN) [1] and Convolutional Neural Network (CNN) has demonstrated remarkable success in variable tasks, including image synthesis [2], [3], data augmentation [4], [5], style transfer [6], [7], Gait analysis [8] and anomaly detection [9], [10]. The most distinctive feature of GANs is the discriminator $D(x)$ that evaluates the divergence between the generative distribution $p_g(x)$ and the target distribution $p_{data}(x)$ [1], [11]. However, real data have a multimodal distribution [12], [13]; therefore, GANs often train on data distributions, completely missing several modes (called *mode collapse*) [14]. For example, the generative distribution omits one of ten digits for MNIST [15].

The conditional GAN (cGAN) [16] has gained wide attention due to its class-wise controllability [17], [18] and superior performance for complex generation tasks [2], [19], [20]. Among them, class cGAN [21], [22], [23], conditioned on auxiliary label information, typically solves the joint matching problem by decomposing it into two easier sub-problems: marginal matching $\left(p_{data}(x), p_g(x)\right)$ and conditional matching $\left(p_{data}(y|x), p_g(y|x)\right)$ [24].

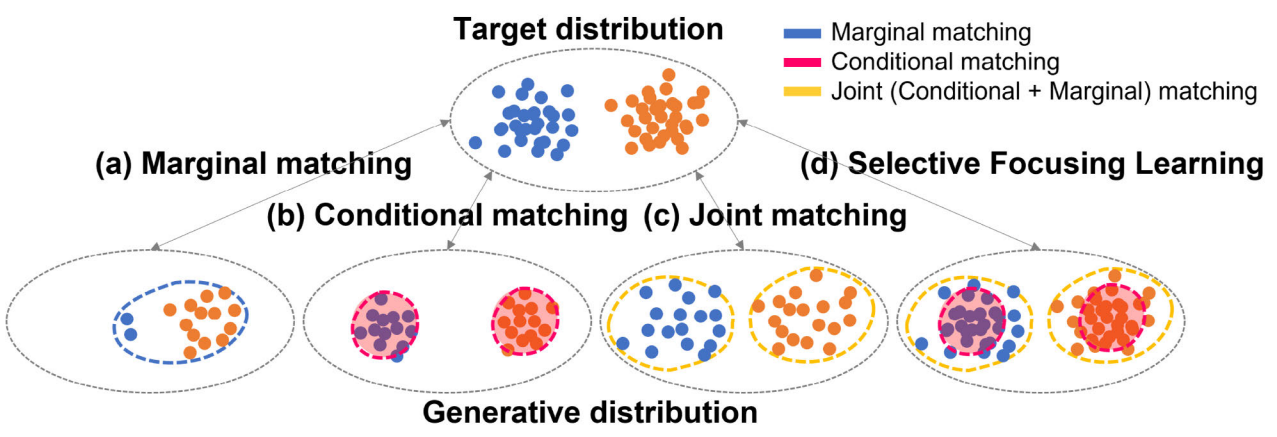

**FIGURE 1. Overview of marginal matching, conditional matching, joint matching, and selective focusing learning.**

The goal of marginal matching is to estimate the generative distribution for the entire target distribution. Since it is similar to the unconditional GAN, the generator focuses on a subset of modes, thereby excluding other parts of the target distribution [see Fig. 1(a)]. On the other hand, conditional matching decomposes the target distribution into smaller sub-distributions using labels and estimates each sub-distribution more easily through the generator. However, the generator tends to focus on high fidelity samples (easy to classify samples) [see Fig. 1(b)]. Applying both marginal and conditional matching for all samples can alleviate the mode collapse issue. However, the joint matching focuses less on high fidelity samples than conditional matching [see Fig. 1(c)]. For simplicity, we denote cGANs as those conditioned on class labels throughout the paper.

This paper proposes a novel training methodology, Selective Focusing Learning (SFL), which enforces the discriminator and generator to learn the easy samples rapidly while maintaining diversity. As illustrated in Fig. 1(d), our key idea is to selectively apply conditional and joint matching for the data in mini-batches. Specifically, we first select the samples with the highest scores when sorted using the conditional term of the discriminator outputs (real and generated samples). Then we optimize the model using the selected samples with only conditional matching and the other samples with joint matching. The precision of the easy sample selection depends on discriminator performance; thus, a proportion applying conditional matching gradually increases as the training step progresses until the certain ratio. Overall, by applying only conditional matching (by freeing the marginal matching) to easy samples, the generator can make samples with high fidelity. By applying the joint matching to the remaining samples, diversity can also be maintained.

Recently, many techniques have been proposed to improve GAN training [25], [26], [27], [28], [29]. Top-k training of GANs [26] is a simple modification to the GAN training algorithm which improves performance by throwing away bad samples. Instance selection for GANs [25] analyzes the use of instance selection [30] in the conditional generative setting. The proposed SFL and these techniques share similar spirit in that utilizing 'realistic (or easy)' samples can help GAN training. However, the methodology and direction are entirely different. While both recent techniques remove *bad samples*, SFL focuses on *good samples* maintaining entire samples in training. Specifically, top-k training zeros out the gradient contributions from the 'least realistic' generated samples. Instance selection for the GAN removes low density regions (hard samples) from the data manifold prior to model optimization (the dataset curation step). In contrast, SFL focuses on easy samples to make a strong discriminator and generator by utilizing content-aware optimization of the discriminator (i.e., training the samples with the most common patterns for each class through conditional matching). In addition, instance selection for GANs improves the overall image sample quality in exchange for reduction in diversity. However, SFL learns the easy samples of each class rapidly without sacrificing diversity by applying joint matching to the remaining samples.

An advantage of SFL is the *flexibility* regarding collaboration with other orthogonal studies, which improves GAN training (instance selection [25], top-k [26]) because it only needs a simple modification in the gradient descent step. We demonstrate the compatibility of the proposed method with these techniques in Section IV-E. In addition, despite the remarkable performance of GANs, there is a significant gap in quality and diversity between class-conditional GANs trained on labeled data and unconditional GANs trained without any labels in a fully unsupervised setting. It is because class-conditional GANs alleviate the mode collapse problem by enforcing labels to include all semantic categories. However, there is a limitation that labels are necessary in the training dataset. Recently, a self-conditional GAN, which can train the class-conditional GAN without the label through clustering technique, is recently being actively studied. Therefore, the proposed method can be used not only in class-conditional GAN such as BigGAN or SA-GAN but also in the unconditional GAN such as StyleGAN or PG-GAN with self-conditional methodology.

Our contributions can be summarized as follows:

- The proposed method can be effectively applied to any cGAN variants with negligible additional time complexity and requires only a few lines to implement.
- We conducted experiments on ImageNet ($64 \times 64$ and $128 \times 128$), CIFAR-10, CIFAR-100 datasets, and Mixture of Gaussian, noisy label settings to demonstrate that the proposed method can substantially improve all indicators.
- The proposed SFL is flexible for the collaboration with other orthogonal studies (Instance Selection [25] and Top-k [26]) because it only needs a simple modification in the gradient descent step.

## II. RELATED WORK

### A. REWEIGHTING SAMPLING

The discriminators learned in GANs can be utilized to reweight generated samples [31]. This includes methods like rejection sampling [27], importance sampling [31], [32], and Markov chain Monte Carlo [33], which are typically applied after the training is complete. Instead of filtering samples post-training, some approaches integrate this process into the training itself. For instance, Latent Optimization for Generative Adversarial Networks (LOGAN) [29] refines latent samples in each iteration, though this requires an additional forward and backward pass. The top-k training method for GANs [26] demonstrates that gradients from low-quality generated samples can mislead the model away from the nearest mode. Therefore, ignoring the gradients from the worst samples during each training iteration can enhance the quality of the generated outputs.

### B. CURRICULUM LEARNING

Inspired by the human behavioral perspective, curriculum learning is a learning paradigm that starts learning with easier examples, and gradually takes more complex examples [34], and it is similar to the principle of human teaching [35]. Curriculum learning has been well studied in computer vision [36], [37], natural language processing [38], reinforcement learning [39], [40], and multitask learning [41]. The self-paced learning algorithm [42] that incorporates curriculum learning into the model optimization was proposed to measure easiness accurately. This method defines the ease measured by the loss value for each sample by adding a regularization term. Many studies have further adopted self-paced learning in their tasks to avoid becoming stuck in bad local minima and improve the generalization of their models [43], [44], [45]. The proposed SFL is similar to self-paced learning in that it selects and learns samples that are easy to classify to make a strong discriminator and generator. However, our method has the difference that it utilizes all samples in the mini-batch during training, but uses a different match for each sample.

## III. SELECTIVE FOCUSING LEARNING

In this section, we first observe the effect of marginal and conditional matching. Then, based on our observation, we propose a new learning method that enforces the discriminator and generator to learn the easy samples rapidly while maintaining diversity.

### A. BACKGROUND

Given a pair of data $x$ and label $y$, $\{x_i, y_i\}_{i=1}^{n} \subseteq \mathcal{X} \times \mathcal{Y}$ sampled from the joint distribution $(x_i, y_i) \sim p_{data}(x, y)$, the goal of the cGAN is to estimate a conditional distribution $p_{data}(x|y)$ by approximating a generative distribution $p_g(x|y)$. We let $p_g(x|y)$ denote the conditional distribution specified by a generator function $G : (z, y) \longrightarrow x$ that maps a pair of a latent $z$ and a label $y$ to real data $x$. Instead of directly modeling $p_g(x|y)$, cGAN trains a $G(z, y)$ to minimize the Jensen-Shannon Divergence (JSD) between $p_{data}(x, y)$ and $p_g(x, y)$:

$$\min_G \max_D \underset{(x,y)\sim p_{data}(x,y)}{\mathbb{E}}[\log D(x, y)] + \underset{z\sim p_z, y\sim p_y}{\mathbb{E}}[\log(1 - D(G(z, y), y))], \quad (1)$$

where $D$ is a discriminator and $y$ is the class label, *i.e.*, $\mathcal{Y} = \{1, \ldots, K\}$.

To improve the image generation performance, typical cGANs [21], [22], [23] solve the joint distribution matching problem by decomposing two easier sub-problems: marginal and conditional matchings [24]. In other words, $D(x, y)$ can be decomposed into a sum of two log likelihood ratios:

$$D(x, y) = \underbrace{\log \frac{p_{data}(y|x)}{p_g(y|x)}}_{\text{conditional } D(y|x):=D_c} + \underbrace{\log \frac{p_{data}(x)}{p_g(x)}}_{\text{marginal } D(x):=D_m}, \quad (2)$$

where $D(y|x) := D_c$ is conditional matching and $D(x) := D_m$ is marginal matching.

### B. SELECTIVE CONDITIONAL MATCHING

We propose SFL: a simple modification for the GAN training procedure to focus training on samples that are easy to classify. Our key idea is to utilize conditional matching characteristics for typical cGAN training through subset selection methodology. The insight of our algorithm is simple. The goal of typical cGAN training is to estimate the joint distribution by approximating a target distribution. However, because of insufficient training data or multi-modality of real data, it is hard to estimate the joint distribution stably. Our simple tweaking algorithm selects the easy samples using the discriminator scores and trains selected samples using only conditional matching. This tweaking helps to obtain high fidelity for easy samples during the cGAN training. In addition, by applying the joint matching rather than conditional matching to the remaining samples, diversity can be maintained as in conventional cGAN. In the next section, we elaborate on why this technique is effective.

### C. INSIGHT ON SFL: EMPIRICAL ANALYSIS

We first checked the effect of marginal, conditional, and joint matching of the cGAN through toy experiments. Then, we verified the effectiveness of SFL when the estimation of the joint distribution is unstable.

#### 1) MARGINAL, CONDITIONAL, AND JOINT MATCHINGS

Following [25], we conducted on the ImageNet dataset with a resolution of 64 × 64 using the SA-GAN [46]. To observe the results of the initial training, we only trained 100k out of 500k iterations. Fig. 2 shows the results of each matching (marginal only, conditional only, joint, and the proposed SFL). In Fig. 3 (a) and (b), we describe the quantitative results using the Inception Scores (IS) per samples (fidelity), Recall (diversity) and the Fréchet Inception Distance (FID) (both fidelity and diversity). Note that samples with high IS mean that they have high confidence (are easy to classify) with the pretrained ImageNet classifier. When the network was trained using only marginal matching, as also reported in [14], the generated samples had low IS and Recall values because of mode collapse [see Fig. 3(a)]. When joint matching (both marginal and conditional matching) is applied, the

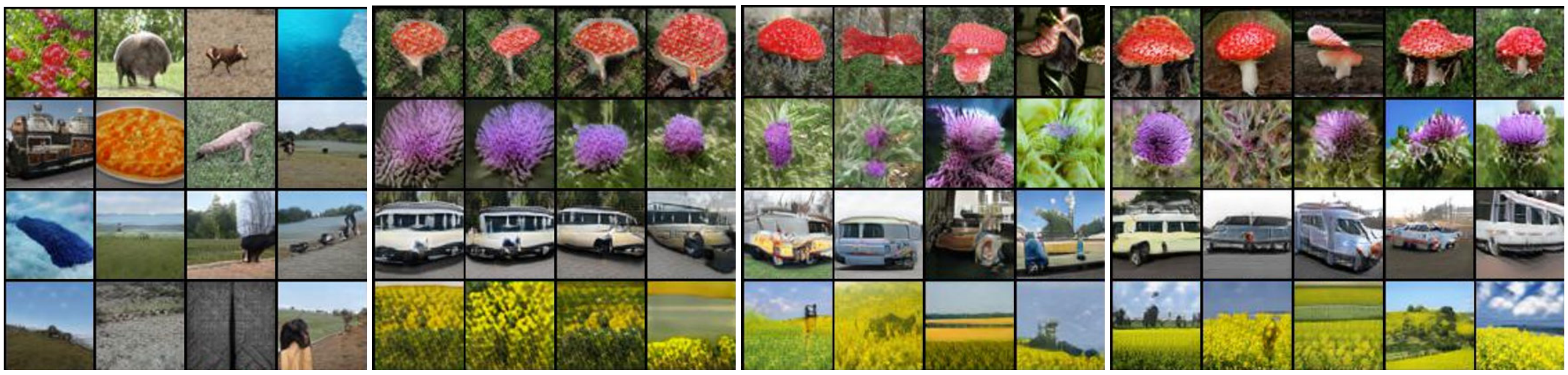

(a) Marginal matching (b) Conditional matching (c) Joint matching (d) SFL

**FIGURE 2.** Example samples generated by conventional methods of marginal matching, conditional matching, joint matching, and selective focusing learning.

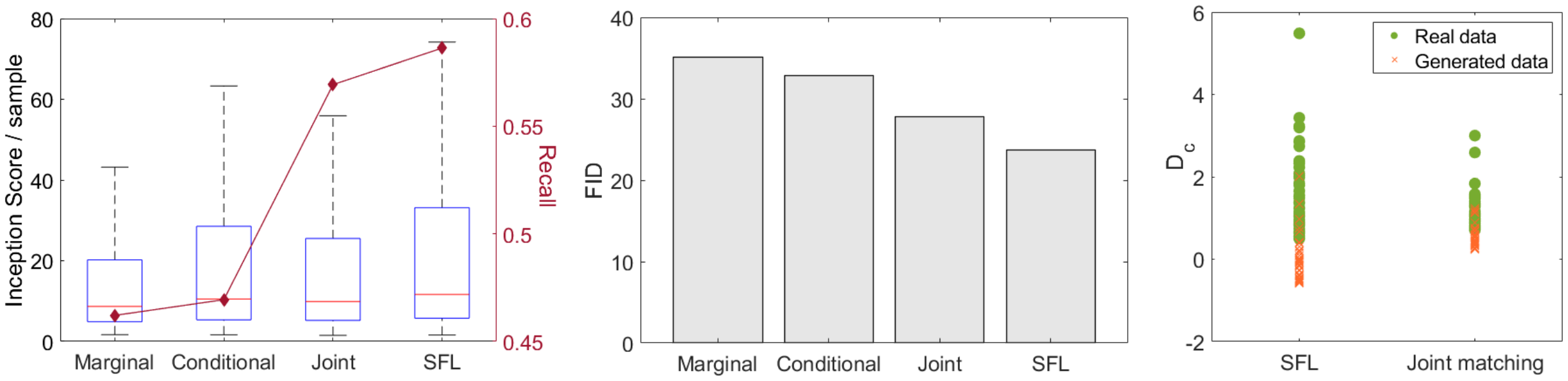

(a) IS per Sample / Recall (b) FID (c) $D_c$ scores

**FIGURE 3.** Quantitative results for each matching trained on ImageNet (64 × 64) with SA-GAN. The SFL has a high variant of the discriminator score compared to joint matching because of content-aware optimization.

**TABLE 1.** Comparison for fine and coarse label datasets. The evaluation indicators P, R, D, and C mean Precision, Recall, Diversity, and Coverage, respectively.

| Label | Method | IS ↑ | FID ↓ | P ↑ | R ↑ | D ↑ | C ↑ |
|---|---|---|---|---|---|---|---|
| Fine | baseline | 9.43 | 8.65 | 0.76 | 0.62 | **0.97** | 0.84 |
| (100 classes) | SFL | **9.60** | **8.15** | **0.77** | **0.64** | **0.97** | **0.87** |
| Coarse | baseline | 8.53 | 10.87 | **0.76** | 0.62 | 0.87 | 0.78 |
| (20 classes) | SFL | **8.90** | **9.61** | **0.76** | **0.63** | **0.92** | **0.82** |

Recall increased significantly, and the IS per sample also increased compared to marginal matching because joint matching can alleviate mode collapse [47]. When only conditional matching was applied, although it has low diversity, samples with high IS were generated compared to joint matching.

Why does this phenomenon happen? In classification tasks, deep neural network (DNN) optimization is content-aware, taking advantage of patterns shared by multiple training examples [48]. In other words, the DNNs learn easy samples with simple patterns first. Since the discriminator of the cGAN plays the role of a classifier, conditional matching generates samples that are easy to classify (samples with high IS) in the initial training stage [see Fig. 2(b)] compared to joint matching [see Fig. 2(c)]. Previous studies [22], [23], [49] also reported that strong classifier leads generators to learn samples that are easy to classify. Whereas they attempt to alleviate this property, the proposed SFL uses this property intuitively to generate easy samples rapidly while maintaining diversity. Applying conditional matching to easy samples and joint matching to the remaining samples can simultaneously achieve the advantages of conditional matching for samples with high IS and joint matching for maintaining high Recall[1] [see Fig. 3(a)]. In addition, by liberating easy samples from marginal matching, this method can accelerate the content-aware optimization of the discriminator compared to joint matching [see Fig. 3(c); increasing variance of conditional term of discriminator output]. As a result, the proposed SFL can generate various high-quality images [see Fig. 2(d)] and achieve the lowest FID score [see Fig. 3(b)].

### 2) FINE VS COARSE LABELS

Recall that the goal of the proposed SFL is also the same as conventional cGAN's in that it estimates a joint distribution by approximating a target distribution. In addition, typical cGANs solve the joint distribution matching problem by decomposing two easier sub-problems: marginal matching and conditional matching. Therefore, conditional matching is also a sub-problem of joint distribution matching. However,

[1]In our toy experiments, Recall of the SFL is also slightly increased compared to joint matching. It is presumed that the discriminator was accelerated through conditional matching in initial training. After training is finished (500k iterations), the joint matching and the SFL have similar Recall (Section IV-A).

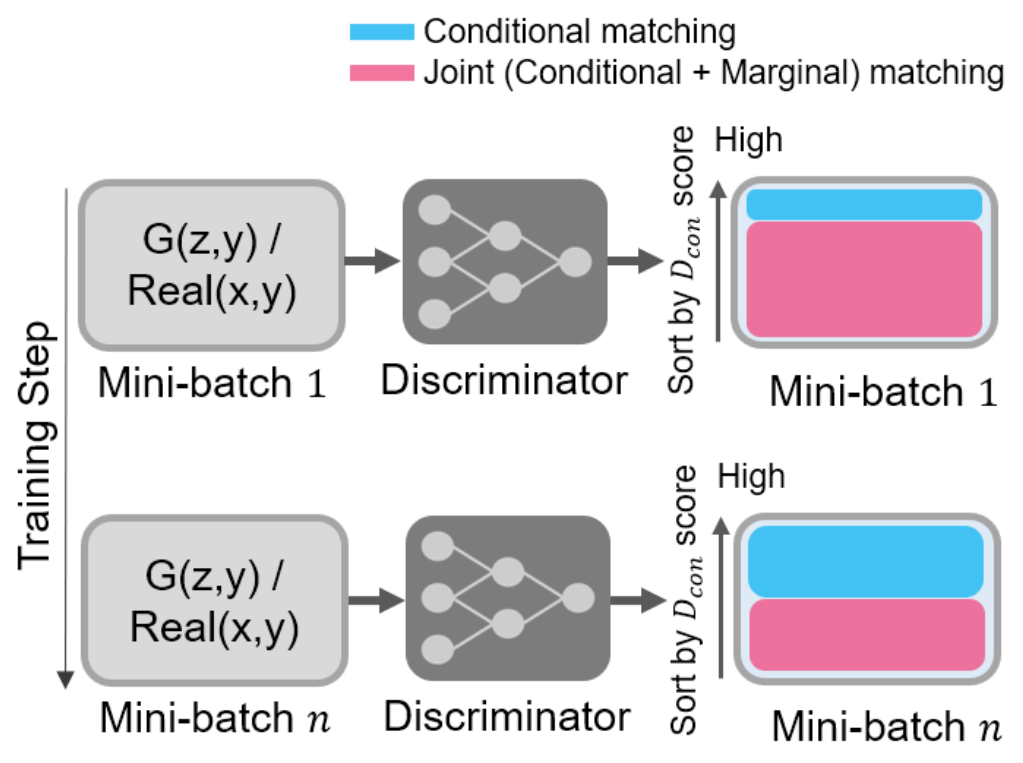

**FIGURE 4. Update procedure of the discriminator parameters on a mini-batch of real and generated samples.**

---

**Algorithm 1** Selective Focus Learning

---

**Input**: $\theta_D$, $\theta_G$, epoch $E_k$ and $E_{max}$, random latent vectors $z$, batch size $B$, conditional loss $\mathcal{L}_{D_c}$, total loss $\mathcal{L}_D$, decay factor $\gamma$, maximum focusing rate $\nu$

**for** e = 1, 2,..., $E_{max}$ **do**

  $F = \min(1-\gamma^e, \nu)$, $k = \lfloor B * F \rfloor$

  **for** $n = 1, \ldots, N_{max}$ **do**

    **Fetch** $\mathcal{X} = \{(x_i, y_i) \sim p_{data}(x, y), i = 1, \ldots, B\}$

    **Fetch** $\mathcal{Z} = \big\{(G(z_i, y_i), y_i) \sim p_g(x, y), i = 1, \ldots, B\big\}$

    **Obtain** $\mathcal{X}_{D_c} = \max_k D_c(\mathcal{X})$, $\mathcal{Z}_{D_c} = \max_k D_c(\mathcal{Z})$

    # $\mathcal{X}_{D_c}$ = DescentSort($D_c(\mathcal{X})$)[:$k$], $\mathcal{Z}_{D_c}$ = DescentSort($D_c(\mathcal{Z})$)[:$k$]

    **Obtain** $\mathcal{X}_D = \overline{\max}_k D_c(\mathcal{X})$, $\mathcal{Z}_D = \overline{\max}_k D_c(\mathcal{Z})$

    # $\mathcal{X}_D$ = DescentSort($D_c(\mathcal{X})$)[$k$:], $\mathcal{Z}_D$ = DescentSort($D_c(\mathcal{Z})$)[$k$:]

    **Update** $\theta_D \leftarrow \theta_D + (\sum_{\mathcal{X}_{D_c}, \mathcal{Z}_{D_c}} \nabla_{\theta_D} \mathcal{L}_{D_c} + \sum_{\mathcal{X}_D, \mathcal{Z}_D} \nabla_{\theta_D} \mathcal{L}_D)$

    **Update** $\theta_G \leftarrow \theta_G - (\sum_{\mathcal{Z}_{D_c}} \nabla_{\theta_G} \mathcal{L}_{D_c} + \sum_{\mathcal{Z}_D} \nabla_{\theta_G} \mathcal{L}_D)$

  **end for**

**end for**

---

in general, cGAN with insufficient training data and coarse label are difficult to accurately estimate the joint distribution. In these cases we guess that the proposed SFL can help the joint matching more stably through performing conditional matching on selected easy samples.

To verify this assumption, we exploited CIFAR-100 dataset which is either categorized into one hundred "fine" classes or twenty "coarse" classes. In Table 1, the proposed SFL can improve the overall performance for both fine and coarse labels. Especially, the proposed SFL was more effective in cases with coarse label than those with fine label (FID improvement (fine vs coarse): 0.5 vs 1.26). This means that when joint matching is unstable, conditional matching for easy samples can significantly help the estimation of high fidelity samples and induce stable training of cGAN. Next, we will describe the algorithm in detail.

### *D. ALGORITHM DESCRIPTION*

As in Fig. 4, when we update the discriminator parameters on a mini-batch of real and generated samples, we applied conditional matching to the elements with the highest scores on the conditional term of the discriminator output and applied joint matching to the remaining elements. Likewise, generator parameters were updated by applying conditional matching to generated samples with the highest scores on the conditional term of the discriminator output and applying joint matching to the remaining elements. When we denote the largest $k$ elements from a set $A$ as $\max_k\{A\}$, the remaining elements as $\overline{\max}_k\{A\}$, $\mathcal{X} = \{(x_i, y_i) \sim p_{data}(x, y), i = 1, \ldots, B\}$, and $\mathcal{Z} = \big\{(G(z_i, y_i), y_i) \sim p_g(x, y), i = 1, \ldots, B\big\}$, we can modify the update step of the discriminator and generator as follows:

$$\theta_D = \theta_D + \alpha_D \left\{ \sum_{\mathcal{X}_{D_c}, \mathcal{Z}_{D_c}} \nabla_{\theta_D} \mathcal{L}_{D_c} + \sum_{\mathcal{X}_D, \mathcal{Z}_D} \nabla_{\theta_D} \mathcal{L}_D \right\}, \quad (3)$$

$$\theta_G = \theta_G - \alpha_G \left\{ \sum_{\mathcal{Z}_{D_c}} \nabla_{\theta_G} \mathcal{L}_{D_c} + \sum_{\mathcal{Z}_D} \nabla_{\theta_G} \mathcal{L}_D \right\}, \quad (4)$$

where

$$\mathcal{X}_{D_c} = \max_k D_c(\mathcal{X}), \quad \mathcal{Z}_{D_c} = \max_k D_c(\mathcal{Z}), \quad (5)$$

$$\mathcal{X}_D = \overline{\max}_k D_c(\mathcal{X}), \quad \mathcal{Z}_D = \overline{\max}_k D_c(\mathcal{Z}), \quad (6)$$

$\mathcal{L}_{D_c}$ and $D_c(\cdot)$ are the conditional term of the loss and discriminator output, and $\mathcal{L}_D$ and $D(\cdot)$ is the total loss and discriminator output, respectively. By performing the SFL on the discriminator predictions, we enforce the generator to learn class-dependent samples while maintaining diversity. The overall procedure of SFL is described in Algorithm 1. The proposed method is easy to implement with few lines (blue comments indicate pseudo-code).

### *E. EXTRACTING CONDITIONAL TERM FROM JOINT DISTRIBUTION*

Conditional term can be expressed in various forms depending on the discriminator type. This paper focuses on evaluating the projection discriminator [22] used as the baseline of the most recent cGANs [2], [29], [46], [50], [51], [52], [53], [54]. Our main idea applies to most types of discriminators of cGANs in which marginal and conditional terms can be divided [21], [55], [56].

The output of the projection discriminator can be represented by a sum of two parametric functions as follows:

$$D(x, y) = D_c + D_m := \mathbf{y}^{\mathrm{T}} V \phi(x; \theta_\Phi) + \psi(\phi(x; \theta_\Phi); \theta_\Psi), \quad (7)$$

where $V$ is the embedding matrix of $\mathbf{y}$, $\phi(\cdot; \theta_\Phi)$ is a vector output function of $x$, and $\psi(\cdot, \theta_\Psi)$ is a scalar function of the same $\phi(x; \theta_\Phi)$ that appears in the first term. The learned parameters $\theta = \{V, \theta_\Phi, \theta_\Psi\}$ are trained to optimize the adversarial loss. Among the two parametric functions, we can simply assume $D_c$ as follows:

$$D_c \approx \tilde{D}_c := \mathbf{y}^{\mathrm{T}} V \phi(x; \theta_\Phi). \quad (8)$$

To derive the exact $D_c$, we would like to elaborate on how [22] can arrive at the two parametric forms. If $y$ is a categorical variable taking a value in $\{1, \cdots, C\}$ and $p_{data}(y|x)$ is obtained using the softmax function, log $p_{data}(y = c|x)$ is represented by the following:

$$\log p_{data}(y = c|x) := (v_c^p)^{\mathrm{T}}\phi(x; \theta_\Phi) - \log Z^p(\phi(x; \theta_\Phi)), \tag{9}$$

where $Z^p(\phi(x; \theta_\Phi)) := \left(\sum_{j=1}^{C} \exp\left((v_j^p)^{\mathrm{T}}\phi(x; \theta_\Phi)\right)\right)$ is the normalization constant and is input into the final layer of the network model. If we parametrize the target distribution $p_g(y = c|x)$ in this form with the same choice of $\phi$, the log likelihood ratio $D(y|x)$ takes the following form:

$$\begin{aligned}\log \frac{p_{data}(y = c|x)}{p_g(y = c|x)} &= (v_c^p - v_c^g)^{\mathrm{T}}\phi(x; \theta_\Phi) \\ &\quad - (\log Z^p(\phi(x; \theta_\Phi)) - \log Z^g(\phi(x; \theta_\Phi))). \end{aligned} \tag{10}$$

Then, if $\mathbf{y}$ denotes a one-hot vector of the label $y$ and $V^p$ and $V^g$ denote the embedding matrices consisting of row vectors $v_c^p$ and $v_c^g$, we can rewrite the above equation as follows:

$$\begin{aligned} D_c := \mathbf{y}^{\mathrm{T}}(V^p - V^g)\phi(x; \theta_\Phi) \\ - \underbrace{(\log Z^p(\phi(x; \theta_\Phi)) - \log Z^g(\phi(x; \theta_\Phi)))}_{\text{normalization constant}}. \end{aligned} \tag{11}$$

For efficient computation, the original projection discriminator [22] integrates $(V^p - V^g)$ into a single embedding matrix $V$ because it can put the normalization constant $(\log Z^p(\phi(x; \theta_\Phi)) - \log Z^g(\phi(x; \theta_\Phi)))$ and marginal term $D_m$ together into one expression $\psi(\phi(x; \theta_\Phi); \theta_\Psi)$. However, because SFL exploits only the conditional term to focus on easy samples, the normalization constant should be separated from the marginal term, and two embedding matrices are necessary. In Section IV-A, we demonstrate that $\tilde{D_c}$ (Approx. SFL) has performance similar to $D_c$ (exact SFL) with smaller computational overhead.

### F. FOCUSING RATE

In the early stages of training, the discriminator may not be a reliable scoring function for self-diagnosing the generator. To help the network estimate class-dependent distribution well, we need to increase the focusing rate $F$ quickly at the initial epochs through a concave function which has zero initial point (see Algorithm 1; line 5). We set $F = \min(1 - \gamma^e, \nu)$ and set $\gamma$ as a convex function as follows: $\gamma = (1 - \nu)^{(1/E_{max})}$, where $e$ and $E_{max}$ are the current and maximum epoch, respectively, and $\nu$ is the maximum focusing rate. At the end of training, $e$ will become $E_{max}$, so, $\gamma^e = \gamma^{E_{max}} = (1 - \nu)$ and the focusing rate $F$ also will become $F = \nu$. In Section IV-A, the effect of $\nu$ is demonstrated empirically.

### G. SFL+: GUIDANCE WITH THE PRETRAINED MODEL

Before the GAN model training occurs, we can predict in advance which samples are easy to classify for the real data using the pretrained classification model. This information can guide the SFL better than unstable discriminator output without additional computational cost. However, because scores are obtained for each class, the score difference between inter-class samples is meaningless. Therefore, instead of using the scores directly, we exploited the *ranking* of samples per class. The notion of *ranking* is formalized as follows.

*Definition 1:* Let $\mathcal{X} = \{x_1, \cdots, x_N\}$ be a set with cardinality $N$. Then, the ranking operator, $\kappa(\cdot)$, takes the elements of $\mathcal{X}$ as the input and output of the indices $\pi(1), \cdots, \pi(N)$ satisfying $x_{\pi(1)} \leq \cdots \leq x_{\pi(N)}$ such that

$$[\pi(1), \cdots, \pi(N)] = \kappa(\mathcal{X}). \tag{12}$$

We define the *ranking*-based SFL as SFL+. The *ranking* is only exploited for the real sample selection, and the remaining steps (including generated samples selection) of SFL+ are the same as those for SFL. In Section IV-A, we describe the effect of SFL+ with the pretrained classification model.

## IV. EXPERIMENTS

In this section, we review evaluation metrics and analyze the impact of SFL for cGANs. We used a variety of evaluation metrics to diagnose the effect of SFL, including the (*i*) *IS* [15], (*ii*) *FID* [57], (*iii*) *Precision and Recall* [58], and (*iv*) *Density and Coverage* [59]. Because the FID cannot be used to analyze fidelity and diversity separately, we also used precision, recall, density, and coverage.

### A. IMAGENET 64 × 64

ImageNet [60] is a large-scale image dataset consisting of over 1.2 million images from 1,000 different classes. To verify the effectiveness of SFL reliably, we conducted all experiments using this benchmark at a resolution of 64 × 64 for 500k iterations. We use single GPU (RTX 2080ti) on ImageNet 64 × 64. For all experiments except for the hyper-parameters under consideration, we set the maximum FR $\nu$ to 50% ($\gamma = (1-\nu)^{(1/E_{max})}$). The remaining parameters are as follows:

SN-GAN: $bs = 64$, $ch = 64$, $G_attn = 0$, $D_attn = 0$, $G_lr = 2e^{-4}$, $D_lr = 2e^{-4}$, $G_step = 1$, $D_step = 5$, and $num_iters = 500000$.

SA-GAN: $bs = 128$, $ch = 32$, $G_attn = 32$, $D_attn = 32$, $G_lr = 1e^{-4}$, $D_lr = 4e^{-4}$, $G_step = 1$, $D_step = 1$, and $num_iters = 500000$.

BigGAN: $bs = 128$, $ch = 64$, $G_attn = 64$, $D_attn = 64$, $dim_z = 120$, $shared_dim = 128$, $G_lr = 1e^{-4}$, $D_lr = 4e^{-4}$, $G_step = 1$, $D_step = 1$, and $num_iters = 500000$.

#### 1) QUANTITATIVE RESULTS

In Table 2, we list the performance of various SFLs. "Approx." uses a conditional term as the $\tilde{D_c}$, and "Exact" uses a conditional term as the $D_c$ as described in Section III-E. In addition, SFL uses a discriminator-based selection of real data, and SFL+ uses a *ranking*-based selection of real data through the pretrained classification model, as described in

**TABLE 2.** Performance of various SFLs with SA-GAN in ImageNet 64 × 64.

| Method | IS ↑ | FID ↓ | P ↑ | R ↑ | D ↑ | C ↑ |
|---|---|---|---|---|---|---|
| Baseline | 17.77 | 17.23±0.15 | 0.68 | 0.66 | 0.72 | 0.71 |
| Approx. SFL | 19.11 | 16.20±0.13 | 0.69 | 0.67 | 0.76 | 0.76 |
| Approx. SFL+ | 21.50 | 14.20±0.11 | 0.72 | **0.68** | 0.84 | 0.80 |
| Exact SFL+ | **21.98** | **13.55±0.12** | **0.73** | 0.66 | **0.85** | **0.81** |

**TABLE 3.** Training time comparison before and after adding SFL on RTX 2080ti GPU.

| Method | Baseline | Approx. SFL+ | Exact SFL+ |
|---|---|---|---|
| Training Time | 35h3m | 35h10m | 46h40m |

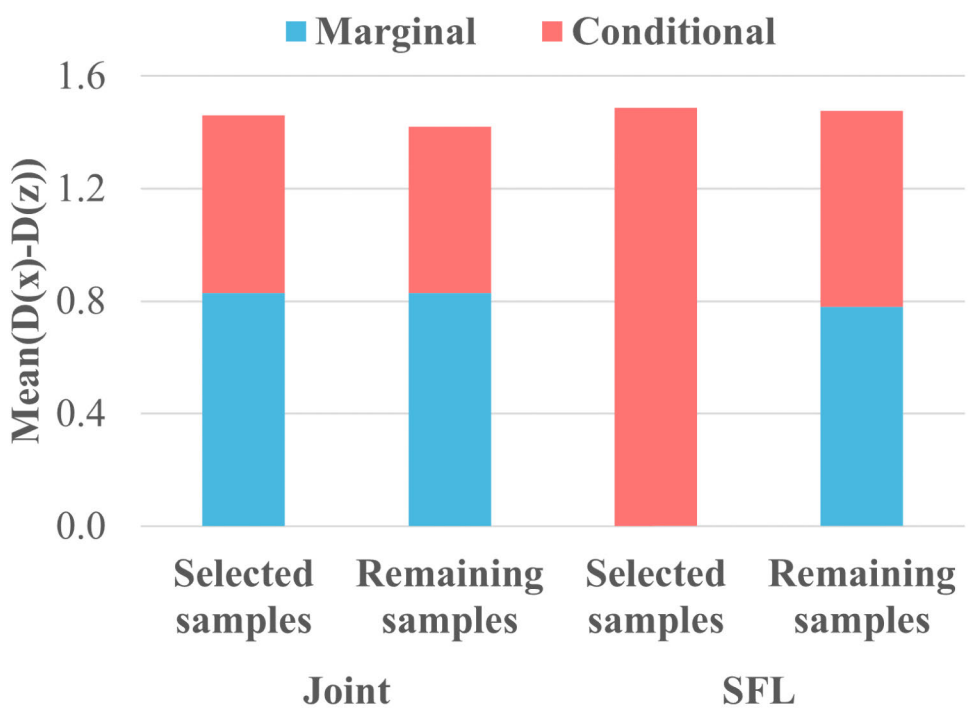

**FIGURE 5.** Comparison of distinguish power between joint matching and SFL.

Section III-G. In the experiments, Approx. SFL and Approx. SFL+ exhibited improved performance compared to the baseline results (without applying SFL) in all metrics. In particular, Approx. SFL+ reduced the FID by 3.03 compared to the baseline, with improvements in both IS (fidelity) and recall (diversity). Lastly, Exact SFL+ achieved slightly better performance except for the recall value compared to the approximated method.

### 2) VISUALIZATION OF THE SFL TRAINING PROCESS

We first visualization of baseline and SFL training process, and then analyzed the effect of applying SFL to real and generated data, respectively. In Fig. 6, when the discriminator was learned by applying SFL only to the real data, the performance of the IS and FID degraded. This is because the discriminator easily wins the minimax game. However, when the discriminator is learned by applying SFL to both the real and generated samples, the fidelity and diversity are improved compared to the baseline. This is because content-aware optimization is accelerated by playing a minimax game using real samples that the discriminator distinguishes well and the generated samples that the generator produces well in terms of conditional matching. This phenomenon is similarly observed in Fig. 3(c) through increasing variance of $D_c$ scores of the proposed SFL. Finally, when SFL is applied to the generator, we can achieve additional performance improvement.

### 3) COMPUTATIONAL COST

To evaluate the computational overhead of SFL, we compared the running time of the baseline SA-GAN [46] and SFL variant on ImageNet datasets in Table 3. After training 500k iterations, Approx. SFL+ and Exact SFL+ took 0.3% and 33.1% more time, respectively, than the baseline. Exact SFL+ took more time than the baseline because this method requires two embedding parts in the discriminator to consider the normalization constants as in (11). Therefore, it makes sense to use Approx. SFL+ in terms of performance and computational cost. Unless specified otherwise, we abbreviate “Approx. SFL” to “SFL” in the remaining experiments.

### 4) DOES SFL WORK AS INTENDED?

To compare the differences between joint matching and SFL in detail, we observed which matching is the focus of the discriminator. For this, we calculated the distinguishing power of the discriminator in the similar way to [61]. Distinguish power corresponding to marginal matching (blue color) is calculated by $Mean(D_m(x, y) - D_m(G(z, y), y))$, and distinguish power corresponding to conditional matching (red color) is calculated by $Mean(D_c(x, y) - D_c(G(z, y), y))$. Overall distinguish power is calculated by $Mean(D(x, y) - D(G(z, y), y))$ where $D(x, y) = D_c(x, y) + D_m(x, y)$. Results are obtained by trained SA-GAN on ImageNet (64 × 64) for 100k out of the total 500k iterations. In Fig. 5, conventional joint matching divides the distinguishing power at a similar rate for the marginal or conditional matching regardless of whether it was selected or not using the conditional term. In contrast, SFL uses all of the distinguish power for conditional matching for selected samples while maintaining joint matching similar to conventional method for the remaining samples. This means that the discriminator and the generator play the minimax game, focusing on conditional matching for the selected samples. In addition, diversity is maintained because the minimax game is also performed through joint matching for the remaining samples.

### 5) QUALITATIVE RESULTS

To verify the effectiveness of enforcing the conditional terms for an easy sample, we randomly generated image samples for a certain class, and sorted the samples using the pre-trained ImageNet classifier (top: easy samples, bottom: hard samples). In Fig. 7, (a) and (b) are fully generated samples with and without SFL+. As illustrated in the red box, SFL+ learns the easy sample well compared to the baseline. Overall, SFL+ generated diverse image samples like the baseline. In (c) and (f), we compared the samples corresponding to the red box for other class and obtained similar results.

### 6) GAN VARIANT ARCHITECTURES

We applied SFL to various sophisticated GANs to demonstrate the effectiveness of the proposed method (Table 4). The FQ-Half[‡] and FQ-Full[‡] were trained using 50 and

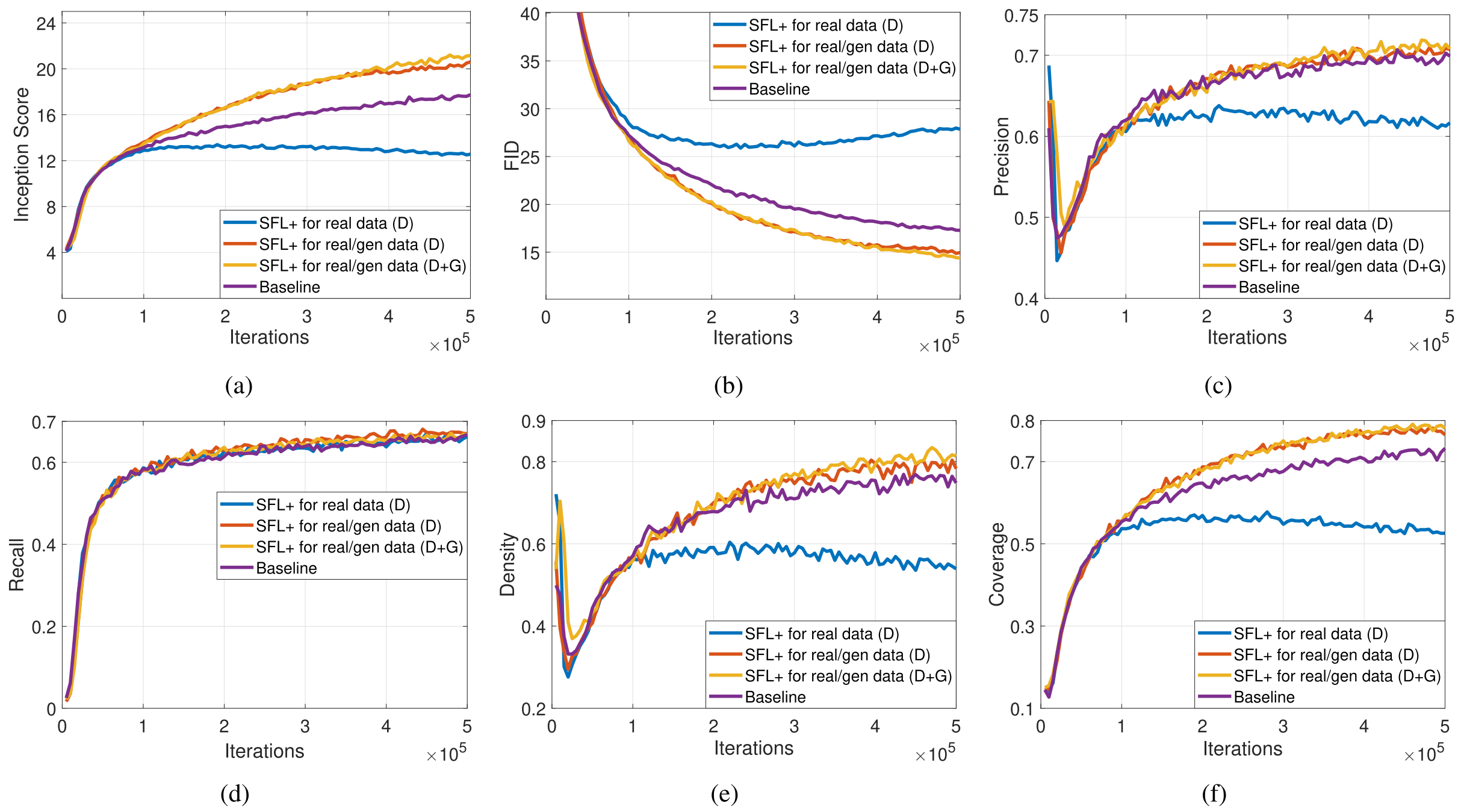

**FIGURE 6.** Visualization of the SFL training process.

**TABLE 4.** The is and FID on the ImageNet dataset for various GAN architectures.

| Metric | Method | SN-GAN Hinge loss | BigGAN Hinge loss | SA-GAN Hinge loss | SA-GAN DC loss |
|---|---|---|---|---|---|
| IS ↑ | Baseline | 10.76 | 20.44 | 17.77 | 16.91 |
| | FQ-Half‡ | - | 21.99 | - | - |
| | FQ-Full‡ | - | 25.96 | - | - |
| | SFL+ | **12.25** | **29.58** | **21.50** | **18.62** |
| FID ↓ | Baseline | 35.68 | 12.79 | 17.23 | 19.44 |
| | FQ-Half‡ | - | 12.62 | - | - |
| | FQ-Full‡ | - | 9.67 | - | - |
| | SFL+ | **32.25** | **8.29** | **14.20** | **17.52** |

100 epochs, respectively, with a 512 batch size, as quoted from FQ-GAN [62]. The rest of the experiments were conducted with a 128 batch size for 50 epochs. In SN-GAN [50] and BigGAN [2] with hinge loss, our method outperformed the baseline by a good margin. In particular, SFL+ in BigGAN outperformed Feature Quantization Full (FQ-Full) [62], the current state-of-the-art model for the task of 64 × 64 ImageNet generation. Despite using 2× fewer epochs and a 4× smaller batch size, our SFL+ achieved a better FID by 1.38. Moreover, SFL+ also outperformed the baseline in the SA-GAN with different losses (DC loss).

### 7) EFFECT OF BATCH SIZE

Recent works [2], [28] suggest that GANs benefit from large batch sizes. To verify the effectiveness of SFL in different batch sizes, we increased batch size $B$ from 64 to 256. SFL+ ($\nu = 50$) is applied to SA-GAN on ImageNet 64 × 64 and is effective for different batch sizes for GAN training. In

**TABLE 5.** Effect of batch size ($B$).

| Metric | Method | SA-GAN $B = 64$ | SA-GAN $B = 128$ | SA-GAN $B = 256$ |
|---|---|---|---|---|
| IS ↑ | Baseline | 15.19 | 17.77 | 18.54 |
| | SFL+ | **19.08** | **21.50** | **22.82** |
| FID ↓ | Baseline | 21.35 | 17.23 | 16.40 |
| | SFL+ | **16.98** | **14.20** | **12.94** |

**TABLE 6.** Effect of the maximum focusing rate.

| Metric | SA-GAN (%) $\nu = 99$ | $\nu = 70$ | $\nu = 50$ | $\nu = 30$ | $\nu = 10$ |
|---|---|---|---|---|---|
| IS ↑ | 18.51 | 21.41 | **21.50** | 20.63 | 18.28 |
| FID ↓ | 17.52 | 14.33 | **14.20** | 14.79 | 16.95 |

Table 5, the baseline performance gradually improved as the batch size increased, and SFL+ outperformed the baseline model by a significant margin regardless of the batch size.

### 8) EFFECT OF MAXIMUM FOCUSING RATE

Our SFL has only one hyper-parameter; the maximum focusing rate $\nu$. SFL+ is applied to an SA-GAN on ImageNet 64 × 64 with different $\nu$. In Table 6, if we use a too large value of $\nu$, it degrades the performance (especially diversity) by enforcing too many samples as conditional matching. Otherwise, using a too small value for $\nu$ degrades the performance because the effectiveness of SFL+ is reduced. In all cases except $\nu = 99$, SFL+ performed better than the baseline (IS: 17.77, FID: 17.23).

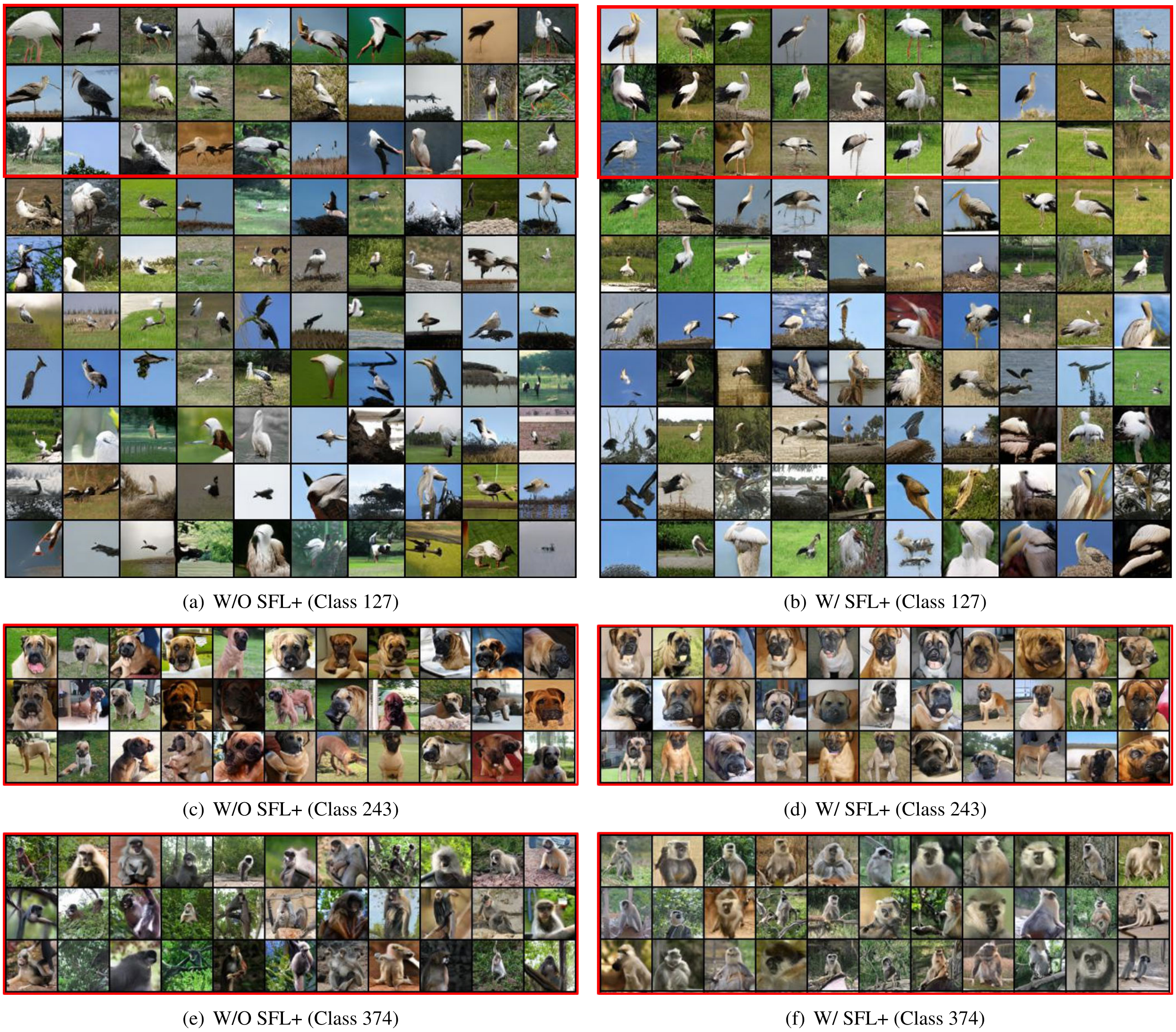

**FIGURE 7. Comparison of the generated samples with and without SFL+ on ImageNet 64 × 64. (a), (b) Full generated samples sorted by easy to hard using the scoring function in [25]. (c)-(f) Samples corresponding to the red box for classes 243, 374.**

### 9) VARIANT EMBEDDING FUNCTIONS

To verify the importance of the embedding function, we compared several different model embeddings that have been trained on different datasets: InceptionV3 [63] trained on ImageNet, ResNet50 [64] trained on Places365 [65], ImageNet, and with SwAV unsupervised pretraining [66], and ResNeXt-101 32 × 8d [67] trained with weak supervision on Instagram 1B [68]. We also compared a randomly initialized InceptionV3 with no pretraining as a random initialization. For all architectures, features were extracted after the global average pooling layer.

From the experiment results, we found that all feature embeddings improved performance of SFL, except for the randomly initialized network. These results show that an embedding function can guide the SFL better than unstable discriminator output. Also, SFL+ with random embedding achieved performance similar to baseline because it is the same as random selection. Interestingly, the Instagram 1B pretrained ResNeXt-101 embedding performed the best overall. This means that the proposed SFL+ is still effective with well-defined feature embedding even if this embedding is not used for evaluation metrics.

### 10) LEARNING WITH NOISY LABEL

We also verify effectiveness of our SFL for noisy vs clean labels. Since ImageNet dataset is clean, following [69], we need to corrupt this dataset manually using the noise transition matrix $Q$, where $Q_i^j = p(\tilde{y} = j|y = i)$, given that noisy is flipped from clean. Among the various noise transition matrices, we used symmetry flipping, which is the case where the true labels of a single class are corrupted by the labels of the other classes for the same ratio. We conducted experiments based on various noise rates:

**TABLE 7.** Comparison of embeddings of different models trained on different datasets.

| Method | Embedding | Pretraining | IS ↑ | FID ↓ | P ↑ | R ↑ |
|---|---|---|---|---|---|---|
| Baseline | - | - | 17.77 | 17.23 | 0.68 | 0.66 |
| SFL | - | - | 19.11 | 16.20 | 0.69 | 0.67 |
| SFL+ | InceptionV3 | ImageNet | **21.50** | 14.20 | **0.72** | **0.68** |
| SFL+ | ResNeXt-101 | Instagram 1B | 21.94 | **13.68** | **0.72** | **0.68** |
| SFL+ | ResNet-50 | ImageNet | 20.74 | 14.45 | 0.71 | 0.67 |
| SFL+ | ResNet-50 | Places365 | 20.13 | 14.85 | **0.72** | 0.67 |
| SFL+ | ResNet-50 | SwAV | 19.96 | 15.26 | 0.70 | 0.67 |
| SFL+ | InceptionV3 | Random init | 17.79 | 17.35 | 0.68 | 0.65 |

**TABLE 8.** Comparison of noisy and clean label datasets.

| Noise rate | Method | IS ↑ | FID ↓ | P ↑ | R ↑ | D ↑ | C ↑ |
|---|---|---|---|---|---|---|---|
| 0% | baseline | 16.06 | 19.91 | 0.67 | **0.65** | 0.68 | 0.67 |
| | SFL+ | **19.00** | **17.32** | **0.70** | **0.65** | **0.77** | **0.75** |
| 20% | baseline | 13.28 | 23.79 | 0.63 | 0.63 | 0.62 | 0.60 |
| | SFL+ | **15.17** | **21.22** | **0.64** | **0.64** | **0.64** | **0.67** |
| 50% | baseline | 12.28 | 28.18 | **0.61** | 0.57 | **0.54** | **0.51** |
| | SFL+ | **12.35** | **26.87** | **0.61** | **0.59** | **0.54** | **0.51** |

**TABLE 9.** Comparison on ImageNet 128 × 128. Baseline‡ and FQ-256$k$‡ were trained for 256$K$ iterations with a 1024 batch size, as quoted in FQ-GAN [62].

| Method | IS ↑ | FID ↓ | P ↑ | R ↑ | D ↑ | C ↑ |
|---|---|---|---|---|---|---|
| Baseline‡ | 63.03 | 14.88 | - | - | - | - |
| FQ-256$k$‡ | 54.36 | 13.77 | - | - | - | - |
| Baseline | 44.29 | 17.55 | 0.75 | **0.65** | 0.90 | 0.75 |
| SFL+ | **72.76** | **10.34** | **0.82** | **0.65** | **1.17** | **0.89** |

$\varepsilon = \{0\%, 20\%, 50\%\}$. In Table 8, the effectiveness of the proposed SFL decreased as the ratio of noise labels increased. This is because the labels were inaccurate at high noise ratio, so the conditional matching was conducted with low accuracy. Nevertheless, the proposed SFL achieved better performance than the baseline for all cases.

### B. IMAGENET 128 × 128

To examine the impact of SFL on high resolution, we conducted all experiments using this benchmark at a resolution of 128 × 128. We use quad GPU (RTX 3090) on ImageNet 128 × 128. Due to the limited hardware resources, compared with the full-version BigGAN, we made the following modifications: $bs = 2048 \rightarrow bs = 256$, $ch = 96 \rightarrow ch = 64$ and $num_iters = 500000$. The remaining parameters are the same as for ImageNet (64 × 64). In Table 9, because we used a smaller batch size (256 vs. 1024) than for FQ-GAN [62], our baseline achieves worse performance than the baseline‡ even when training more iterations (500$k$ vs. 256$k$). Despite using 4× a smaller batch size, the SFL+ achieves the best performance for all metrics.

### C. MIXTURE OF GAUSSIAN

Following [23], we draw samples from 2D Mixture of Gaussian (MoG) distribution with three Gaussian components, labeled as class 1, class 2, and class 3, respectively. The standard deviations of the three components are fixed to $\sigma_1 = 1$, $\sigma_2 = 2$, and $\sigma_3 = 3$. The means were set to $\mu_1 = 0$, $\mu_2 = 3$, and $\mu_3 = 6$. We conducted experiments on a simple MLP network with a projection discriminator [22]. We trained 40 epochs with the batch size of 256. We evaluated the Maximum Mean Discrepancy (MMD) [70], a metric of the distance between the real data and the generated data. Here, models were trained using the cross-entropy loss. The proposed SFL achieved lower MMD values than the baseline for each class distribution. This means that the proposed SFL estimates the center of each class distribution well, and it seems reasonable because the proposed SFL focuses on class-dependent samples. In addition, the proposed SFL achieved a lower MMD value than the baseline for overall distribution.

**TABLE 10.** Comparison on 2D Mixture of Gaussian distribution with three Gaussian components; class 1, class 2, and class 3.

| Data | Baseline | SFL |
|---|---|---|
| Class 1 | 0.041 | **0.006** |
| Class 2 | 1.667 | **0.167** |
| Class 3 | 2.152 | **0.920** |
| Overall distribution | 0.313 | **0.132** |

**TABLE 11.** Comparison using CIFAR-100.

| Method | IS ↑ | FID ↓ | P ↑ | R ↑ | D ↑ | C ↑ |
|---|---|---|---|---|---|---|
| BigGAN | 9.43 | 8.65 | 0.76 | 0.62 | **0.97** | 0.84 |
| SFL BigGAN | **9.60** | **8.15** | **0.77** | **0.64** | **0.97** | **0.87** |

### D. CIFAR-100

We set the maximum FR of $\nu$ to 70%. The remaining parameters are the same as for CIFAR-10. In Table 11, the SFL BigGAN outperformed the baseline BigGAN in all metrics.

### E. CONNECTION TO GAN TRAINING TECHNIQUES

We examined the compatibility of the proposed method with recent training methods for GANs. The experiments were conducted with the datasets that each method mainly used (ImageNet 64 × 64 and CIFAR-10).

**Instance selection for GANs** [25] analyzed instance selection [30] in the conditional generative setting. This method removed low density regions from the data manifold prior to model optimization. It improved the overall image sample quality in exchange for reducing diversity with a small model capacity and training time. By redefining target distribution through instance selection, SFL can be applied to easy target distribution. In this case, the proposed SFL is still effective because there are easy and hard samples in the new target distribution. We applied SFL and SFL+ to the dataset after instance selection. In Table 12, RR and FR are the retention ratio (percentage of remaining dataset after instance selection) and maximum focusing ratio (maximum focusing rate of the remaining datasets), respectively. SFL+ outperformed the baseline for almost all metrics. Because the goal of instance selection is to remove low density regions, it is reasonable to say that the effectiveness of SFL reduced as the retention ratio (RR) reduced. Nevertheless, our SFL+ achieved a value that is 0.36 lower than the best FID in instance selection.

**TABLE 12.** Performance on instance selected ImageNet 64 × 64 with the SA-GAN [46]. ‡ is quoted from [25].

| RR (%) | Method | FR (%) | IS ↑ | FID ↓ | P ↑ | R ↑ | D ↑ | C ↑ |
|---|---|---|---|---|---|---|---|---|
| 80 | Instance Selec.‡ | 0 | 21.62 | 13.17 | 0.74 | **0.65** | 0.87 | 0.79 |
| | SFL | 50 | 24.06 | 12.37 | 0.75 | **0.65** | 0.91 | 0.83 |
| | SFL+ | 50 | **26.67** | **11.29** | **0.76** | **0.65** | **0.97** | **0.85** |
| 60 | Instance Selec.‡ | 0 | 27.95 | 10.35 | 0.78 | **0.63** | 0.99 | 0.87 |
| | SFL | 33 | 30.17 | 9.84 | 0.78 | **0.63** | 1.02 | 0.88 |
| | SFL+ | 33 | **33.38** | **8.79** | **0.80** | **0.63** | **1.12** | **0.90** |
| 40 | Instance Selec.‡ | 0 | 37.10 | 9.07 | 0.81 | **0.60** | 1.12 | 0.90 |
| | SFL | 12.5 | 37.52 | 8.87 | 0.82 | **0.60** | 1.16 | 0.91 |
| | SFL+ | 12.5 | **40.65** | **8.71** | **0.83** | 0.59 | **1.20** | **0.92** |

**TABLE 13.** Comparison to the top-k training of GANs on CIFAR-10. ‡ is quoted from [62].

| Method | IS ↑ | FID ↓ | P ↑ | R ↑ | D ↑ | C ↑ |
|---|---|---|---|---|---|---|
| SN-GAN‡ | 8.22 | 14.26 | - | - | - | - |
| R-MMD-GAN‡ | 8.29 | 16.21 | - | - | - | - |
| BigGAN | 8.43 | 6.45 | **0.76** | 0.65 | 1.01 | 0.88 |
| FQ-BigGAN‡ | 8.48 | 5.59 | - | - | - | - |
| Top-k BigGAN | 8.45 | 6.04 | 0.75 | 0.66 | 0.98 | 0.89 |
| SFL BigGAN | 8.60 | 5.89 | **0.76** | 0.66 | 1.01 | 0.91 |
| Both BigGAN | **8.78** | **5.25** | **0.76** | **0.67** | **1.02** | **0.92** |

**The top-k training of GANs** is a simple modification to the GAN training algorithm, improving performance by removing bad samples [26]. Since SFL also generates bad samples during the training, top-k can improve the performance of SFL. In Table 13, the top-k BigGAN outperformed the baseline BigGAN in all metrics except for precision and density. Further, SFL achieved better performance than top-k, and we can achieve state-of-the-art performance by applying both methods.

## V. CONCLUSION

In this paper, we proposed SFL, which enforces the discriminator and generator to learn easy samples rapidly while maintaining diversity. The proposed method can easily be applied to any cGAN variant and requires only a few lines to implement. The experiment results showed that image quality of easy samples can be significantly improved without sacrificing diversity by the selective focusing on easy samples.

Furthermore, this work provides a unique perspective on a promising field: GAN training. Improvements to image generation results can be adjusted to improve photo editing efficiency and generate more realistic simulations for robot training. Nevertheless, GANs can also jeopardize personal identity by generating fake images of people's faces (e.g., deepfakes). We hope that future work will address this issue through deep fake detection and contribute to the positive social development of the technology.

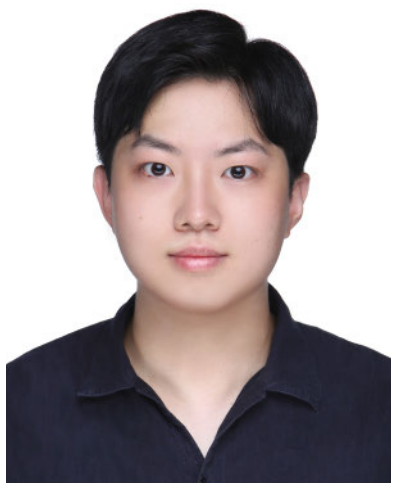

**KYEONGBO KONG** (Member, IEEE) received the B.S. degree in electronics engineering from Sogang University, Seoul, South Korea, in 2015, and the M.S. and Ph.D. degrees in electrical engineering from Pohang University of Science and Technology (POSTECH), Pohang, South Korea, in 2017 and 2020, respectively. From 2020 to 2021, he was a Postdoctoral Fellow with the Department of Electrical Engineering, POSTECH. From 2021 to 2023, he was an Assistant Professor with the Media School, Pukyong National University, Busan. He is currently an Assistant Professor in electrical and electronics engineering with Pusan National University. His current research interests include image analysis and enhancement, video processing, multimedia signal processing, 3D vision, and generative model.

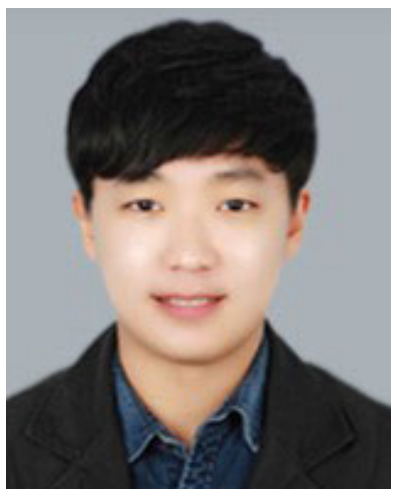

**KYUNGHUN KIM** received the B.S. degree in computer engineering from Yonsei University, in 2019, and the M.S. degree in electrical engineering from Sogang University, Seoul, in 2021. In the NAVER AI Challenge 2020, he ranked second in the ‘‘Spam mail classification’’ in the natural language processing field, the second place in the ‘‘Sound source classification’’ in the image and text processing field, and the third place in the ‘‘Naver shopping review image automatic tagging’’ in the image processing field. He is currently a Research Scientist with NHN Cloud. His current research interests include object detection, generative model, computer vision, and deep learning.

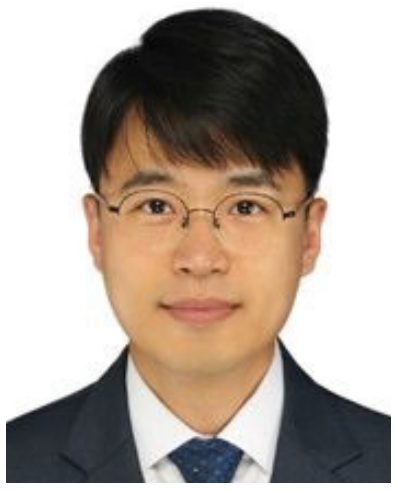

**SUK-JU KANG** (Member, IEEE) received the B.S. degree in electronic engineering from Sogang University, South Korea, in 2006, and the Ph.D. degree in electrical and computer engineering from Pohang University of Science and Technology, in 2011. From 2011 to 2012, he was a Senior Researcher with LG Display, where he was the Project Leader of resolution enhancement and multiview 3D system projects. From 2012 to 2015, he was an Assistant Professor in electrical engineering with Dong-A University, Busan. He is currently a Professor in electronic engineering with Sogang University. His current research interests include image analysis and enhancement, video processing, multimedia signal processing, circuit design for display systems, and deep learning systems. He was a recipient of the IEIE/IEEE Joint Award for Young IT Engineer, in 2019, and the Merck Young Scientist Award, in 2022.